\documentclass[sigconf,nonacm]{acmart}
\renewcommand\footnotetextcopyrightpermission[1]{} 
\copyrightyear{2020}
\acmYear{2020}
\setcopyright{acmcopyright}
\acmDOI{10.1145/1122445.1122456}

\acmConference[MM '20] {28th ACM International Conference on Multimedia}{October 12--16, 2020}{Seattle, WA, USA.}
\acmBooktitle{28th ACM International Conference on Multimedia (MM '20), October 12--16, 2020, Seattle, WA, USA.}
\acmPrice{15.00}
\acmDOI{10.1145/3394171.3413585}
\acmISBN{978-1-4503-7988-5/20/10}

\usepackage{subfigure}
\usepackage{multirow}
\usepackage{xcolor}
\usepackage{mathrsfs}
\usepackage{amsmath}
\usepackage{amsthm}
\begin{document}
\fancyfoot{}
\title{Light Field Super-resolution via Attention-Guided Fusion of Hybrid Lenses}
\thispagestyle{empty}

\author{Jing Jin}
\affiliation{%
  \institution{City University of Hong Kong}}
\email{jingjin25-c@my.cityu.edu.hk}

\author{Junhui Hou}
\authornote{Corresponding author.}
\affiliation{%
  \institution{City University of Hong Kong}}
\email{jh.hou@cityu.edu.hk}

\author{Jie Chen}
\affiliation{%
  \institution{Hong Kong Baptist University}}
\email{chenjie@comp.hkbu.edu.hk}

\author{Sam Kwong}
\affiliation{%
  \institution{City University of Hong Kong}}
\email{cssamk@cityu.edu.hk}

\author{Jingyi Yu}
\affiliation{%
  \institution{Shanghai Tech University}}
\email{yujingyi@shanghaitech.edu.cn}


\begin{abstract}
    This paper explores the problem of reconstructing high-resolution light field (LF) images from hybrid lenses, including a high-resolution camera surrounded by multiple low-resolution cameras. 
    To tackle this challenge, we propose a novel end-to-end learning-based approach, which can comprehensively utilize the specific characteristics of the input from two complementary and parallel perspectives.
    Specifically, one module regresses a spatially consistent intermediate estimation by learning a deep multidimensional and cross-domain feature representation; the other one constructs another intermediate estimation, which maintains the high-frequency textures, by propagating the information of the high-resolution view.
    We finally leverage the advantages of the two intermediate estimations via the learned attention maps, leading to the final high-resolution LF image. Extensive experiments demonstrate the significant superiority of our approach over state-of-the-art ones.
    That is, our method not only improves the PSNR by more than 2 dB, but also preserves the LF structure much better. To the best of our knowledge, this is the first end-to-end deep learning method for reconstructing a high-resolution LF image with a hybrid input.
    We believe our framework could potentially decrease the cost of high-resolution LF data acquisition and also be beneficial to LF data storage and transmission.
    The code is available at \url{https://github.com/jingjin25/LFhybridSR-Fusion}.
\end{abstract}



\maketitle

    \begin{figure}[t]
    \begin{center}
    \includegraphics[width=\linewidth]{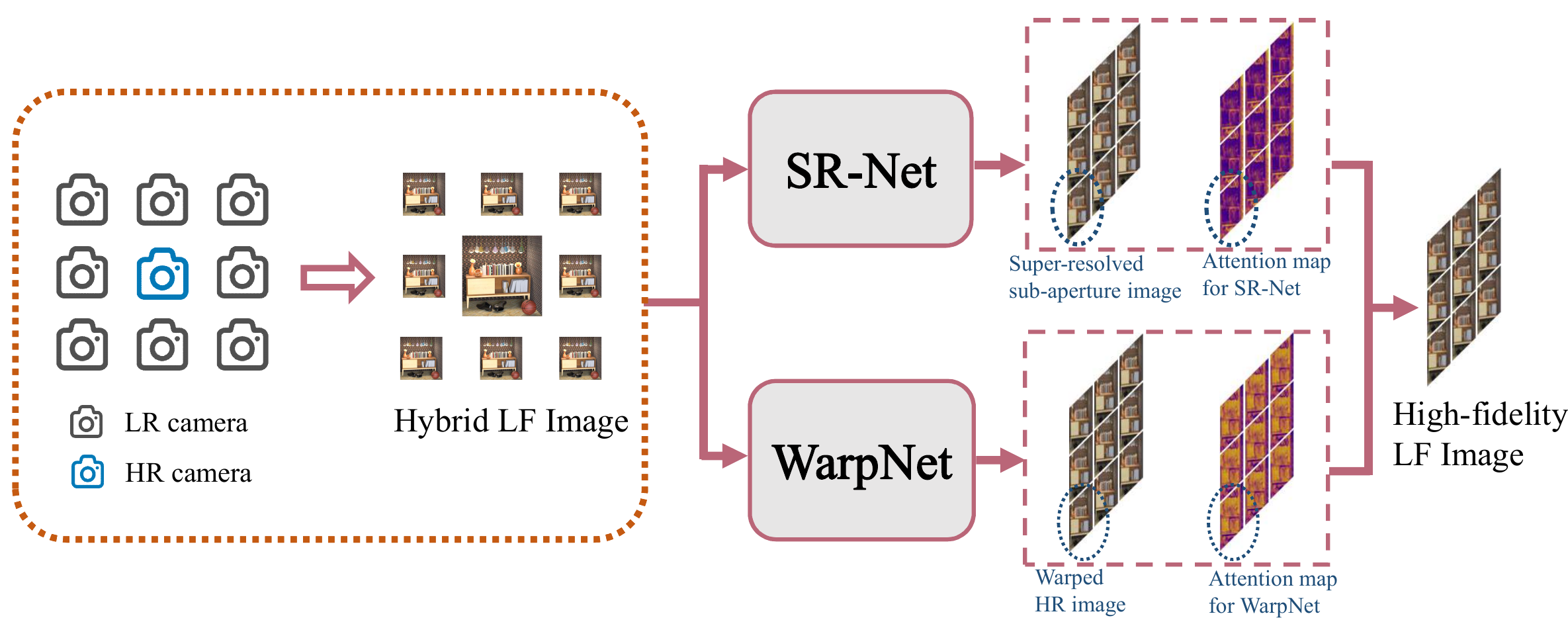}
    \end{center}
      \caption{The illustration of the proposed framework. The hybrid imaging system captures an HR central view and multiple LR views. Two modules are involved to reconstruct the HR LF image, and the predictions of them are fused based on learned attention maps. 
      }
    \label{fig:flowchat}
    \end{figure}

\section{Introduction}

    The light field (LF) describes all light rays through every point along every direction in a free space. 
    An LF image can be interpreted as multiple views observed from viewpoints regularly distributed over a 2-D grid. Therefore, LF images contain not only color information but also geometric structure of the scene in an implicit manner. The rich information enables many applications such as image post-refocusing \cite{lfapp2014refocusing}, material recognition \cite{lfapp2016material}, saliency detection \cite{lfapp2014saliency}, motion deblurring \cite{lfapp2017motion}, 3D reconstruction \cite{lfapp2013scene}, and virtual/augment reality~\cite{lfapp2017vryu}. Recent research also demonstrates that LF is a promising media for virtual/augment reality \cite{lfapp2015vr,lfapp2017vryu}.

    A high-quality LF image can be captured by a densely positioned array of high-resolution (HR) cameras. However, it is neither practical or necessary to do so with so many separate HD units.
    Recent commercialized  LF cameras provide a convenient way to capture LF images. However, the captured LF images always suffer from low spatial resolution due to the limitation of sensor resolution. 
    To overcome this limitation, many methods for reconstructing HR LF images have been proposed~\cite{lfssr2014variational,lfssr2017graph,lfssr2015yoon,lfasr2020aaai,lfasr2020eccv,lfasr2018Yeung,lfhybrid2018dsp,lfhybrid2014iccp,lfhybrid2017ring,lfhybrid2016splitter,lfhybrid2018tci}.
    Among them, LF reconstruction with a hybrid input is a promising way. 
    A hybrid LF imaging system can be built by a sparse grid of low-resolution (LR) image sensors that surround a central HR camera ~\cite{lfhybrid2017ring,lfhybrid2018tci}.
    These heterogeneous sensors simultaneously sample along the angular and spatial dimensions of the LF at different sampling rates, and provide sufficient information for subsequent algorithms to calculate an HR LF.
    The LR views are useful for recording the geometry information of the scene, while the HR central view captures delicate textures and high-frequency information of the scene. To produce an HR LF image, a post-process algorithm is necessary to combine the information of the hybrid input.

    Although multiple algorithms have been proposed to reconstruct an HR LF from the hybrid input \cite{lfhybrid2014iccp,lfhybrid2016splitter,lfhybrid2018tci,lfhybrid2017ring}, they still have limited performance. Generally, these methods comprise several steps that are independently designed, and the final results would be compromised by any inaccuracy of each step. Furthermore, these methods fail to fully describe the complicated relation between 
    the HR central view and the LR side views as well as the one within the high-dimensional LF image.

    We propose a learning-based framework to reconstruct an HR LF image with a hybrid input in an end-to-end manner.
    To the best of our knowledge, this is the first one to study a hybrid algorithm for LF reconstruction using deep learning techniques. 
    The proposed framework produces impressive performance. As illustrated in Figure \ref{fig:flowchat}, our framework achieves the goal with two \textit{complementary} and \textit{parallel} research lines, namely SR-Net and Warp-Net, and the advantages of them are combined via attention-guided fusion.
    The SR-Net up-samples the LR views to the desired resolution by learning a deep representation from both components of the hybrid input. The results of this module are spatially consistent with respect to the scene content, but always blurred, especially when the up-sampling scale is relatively large. In Warp-Net, the HR view is warped to synthesize an HR LF using the disparity maps estimated from the LR views. The predictions by this module inherit the delicate textures and high-frequency information from the HR view, but always have artifacts caused by occlusion or disparity inaccuracy. Observing the complementary behavior between these two modules, we learn a pixel-wise attention map for the output of each module. And, the final HR LF image is obtained by adaptively fusing the two intermediate predictions based on their attention maps, in which only their advantages are collected. 
    
    The rest of this paper is organized as follows. Sec. 2 comprehensively reviews existing methods for LF super-resolution. Sec. 3 presents the proposed attention-guided method. Sec. 4 demonstrates the advantages of the proposed method through extensive experiments. Sec. 5 discusses the benefits of the proposed method in real hybrid systems and potential LF applications. Finally, Sec. 6 concludes this paper.

\section{Related Work}
\textbf{LF Image Super-resolution.}
Single image super-resolution (SISR) is a classical problem in the field of image processing. To solve this ill-posed inverse problem, lots of regularization based methods \cite{sisr2015self,sisr2004ne, sisr2014a+} have been proposed. Recently deep learning based methods \cite{sisr2016srcnn, sisr2016vdsr, sisr2017lapsrn, sisr2018residual,sisr2016perceptual, sisr2017gan} have achieved great success. We refer the reader to \cite{sisr2011survey2, sisr2019survey1} for comprehensive review on SISR.

Different from SISR, LF image super-resolution aims at simultaneously increasing the spatial resolution of all sub-aperture images (SAIs) in an LF image. On top of the target to recover high-frequency details for each SAI, LF super-resolution should also maintain the LF structure. To characterize the relation between SAIs, many methods define a physical model to reconstruct the observed LR SAIs using the desired HR ones. Afterwards, the inverse problem is solved by different priors \cite{lfssr2012gmm,lfssr2014variational,lfssr2014rpca,lfssr2017graph}. These methods always require accurate disparity estimation.

Recent years have witnessed progress on learning-based methods for LF super-resolution. Farrugia \textit{et al.} \cite{lfssr2017subspace} constructed a training set by 3D-stacks of  2-D-patches cropped from different SAIs of paired LF images, and then learned a linear mapping between the subspace of the LR and HR patch stacks.
Yoon \textit{et al.} \cite{lfssr2015yoon} are the first to apply convolutional neural network (CNN) on LF images. However each SAI of an LF image is processed independently in their network, which neglects the angular relationship. Therefore, Yuan \textit{et al.} \cite{lfssr2018combine} proposed to refine the result after separately applying an SISR approach on each SAI.
For the same purpose of keeping the geometric consistency in the reconstructed LF image, Wang \textit{et al.} \cite{lfssr2018recurrent} adopted a recurrent neural network to learn the relations between adjacent SAIs along horizontal and vertical directions. 
To take advantage of the complementary information between SAIs introduced by the LF structure and address the high-dimensionality challenging, Yeung \textit{et al.} \cite{lfssr2018separable} proposed to use 4-D convolution and more efficient spatial-angular separable convolution on LF images, and Jin \textit{et al.} \cite{lfssr2020cvpr} proposed an All-to-One module to fuse the combinatorial geometry embedding between the reference and auxiliary views in the LF image, which achieves the state-of-the-art performance.

\textbf{LF Image Super-resolution with a Hybrid Input}.
LF hybrid imaging system was first proposed by Lu \textit{et al.} \cite{lfhybrid2013microscopy}, in which an HR RGB camera is co-located with a Stack-Hartmann sensor.
Boominathan \textit{et al.} \cite{lfhybrid2014iccp} proposed a patch-based method named \textit{PaSR} to improve the resolution with the hybrid input. Based on \textit{PaSR}, Wang \textit{et al.} \cite{lfhybrid2017ring} improved the performance by iterating between patch-based super-resolution and depth-based synthesis, where the synthesized images were used to update the patch dictionary. The patch-based approaches avoid the need to calibrate and register the DSLR camera and the LF camera. However, the average aggregation causes blurring. Zhao \textit{et al.} \cite{lfhybrid2018tci} proposed a method named \textit{HCSR} to separate the high-frequency details from the HR image and warp them to all SAIs to reconstruct an HR LF image. A similar but simpler task is reference-based SISR. Zheng \textit{et al.} \cite{lfhybrid2018crossnet} designed a network, named \textit{CrossNet}, to align the information of an HR image to the target LR image which is captured from a different viewpoint. Besides spatial super-resolution, the hybrid LF imaging system was also used to generate LF videos \cite{lfhybrid2017video}.

    \begin{figure*}
    \begin{center}
    \includegraphics[width=\textwidth]{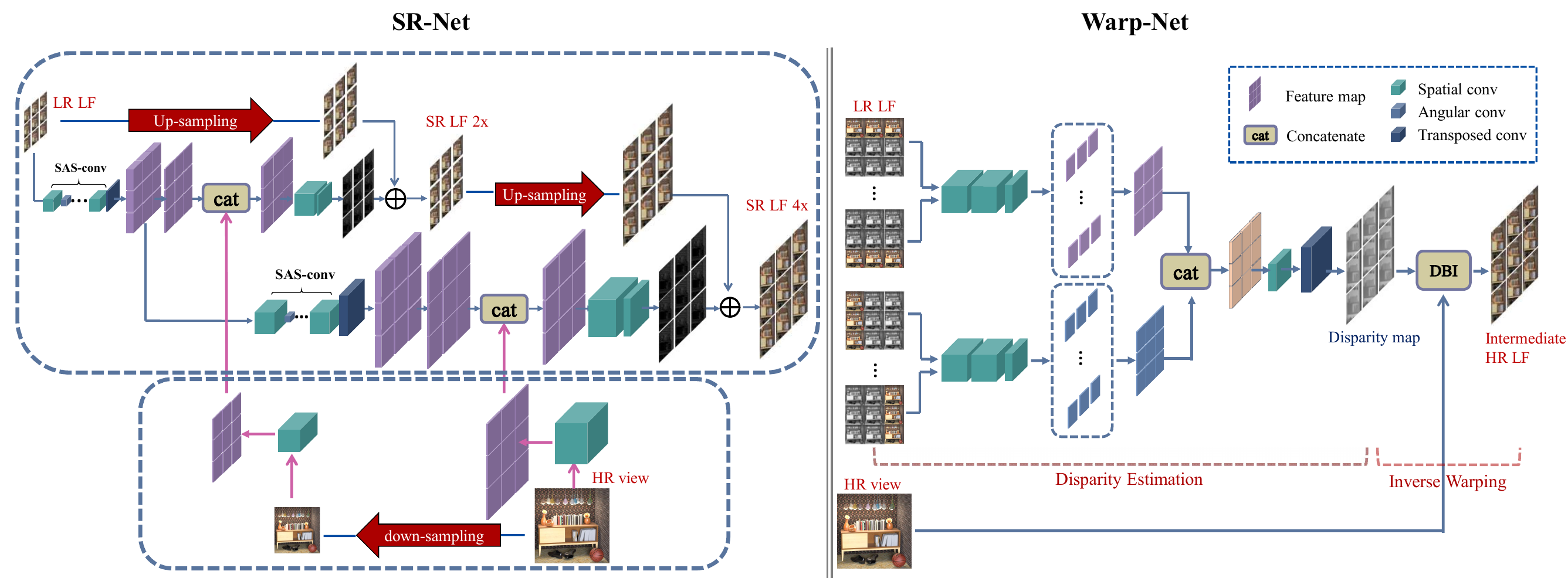}
    \end{center}
      \caption{Illustration of the network architectures of the SR-Net (left) and WarpNet (right). 
      The SR-Net progressively super-resolves the LR LF in a coarse-to-fine manner, and the information from the HR central view is progressively extracted and brought in along an inverse direction.
      Note that the left figure illustrates the SR-Net of 4$\times$ reconstruction, and the model for a larger scale can be realized by cascading more levels.
      The WarpNet first estimates and spatially super-resolves the  4-D disparity maps from the LR LF, and then warps the HR view to synthesis an intermediate HR LF image. \textit{DBI} stands for the Differentiable Bicubic Interpolation. }
    \label{fig:subnets}
    \end{figure*}

\section{The Proposed Approach}
\subsection{Overview}

Let $L^l (x, y, s, t)$ ($1\leq x\leq W$, $1\leq y\leq H$, $1\leq s\leq M$, $1\leq t\leq N$) denote an LR  4-D LF image of spatial resolution $W\times H$ and angular resolution $M\times N $. An HR 4-D LF image to be reconstructed is denoted as $\tilde{L}^h (x', y', s' , t')$ ($1\leq x'\leq \alpha W$, $1\leq y'\leq \alpha H$, $1\leq s'\leq M$, $1\leq t'\leq N$),  where $\alpha >1$ is the up-sampling scale factor, and the corresponding ground-truth one is denoted as $L^h$.
The problem of reconstructing $\tilde{L}^h$ from the hybrid input can be implicitly formulated as
\begin{equation}
      \tilde{L}^h = \mathcal{F}(I^h, L^l_s),
\end{equation}
where $I^h$ denotes the HR central view, and $L^l_s$ denotes the surrounding LR LF views. Note that the complete LR LF $L^l$ can be obtained by down-sampling $I^h$ as the LR central view. Considering the powerful representation ability of deep CNNs, we investigate a deep neural network that can well capture the characteristics of the input to learn such a mapping function $\mathcal{F}$.

To reconstruct $\tilde{L}^h$, the specific properties of the hybrid input $L^l$ and $I^h$ have to be fully explored.  As shown in Figure \ref{fig:flowchat}, our approach consists of two modules, namely SR-Net and Warp-Net. More specifically, by efficiently learning deep representations from both $L^l$ and $I^h$, the SR-Net aims to super-resolve $L^l$ in a progressive manner, (i.e., to equally increase the spatial resolution of all views contained in $L^l$), leading to an intermediate HR LF image and its corresponding attention map. The Warp-Net learns another intermediate HR LF image as well as its attention map, in which $I^h$ is inversely warped with the disparity estimated from $L^l$. Finally, the two intermediate predictions are adaptively fused based on the learned attention maps such that only their respective advantages can be combined into a better output. Note that our approach is trained end-to-end.  In the following, the details of the proposed approach as well as comprehensive analyses are presented step-by-step. 

\subsection{SR-Net}
As depicted in Figure \ref{fig:subnets}, the SR-Net is composed of two branches, i.e., an up-sampling branch used to progressively up-sample $L^l$, and a down-sampling branch used to progressively extract features from $I^h$. The two structurally opposite branches are connected to promote sufficient exploration of the information contained in the hybrid input.

\textbf{The Up-sampling Branch.} The high dimensionality of LF data makes it memory and time consuming for convolutional operations. To address this challenging issue, we adopt a cascaded hourglass structure \cite{sisr2017lapsrn,lfssr2018separable}, which sequentially extract features at the LR level and recursively up-samples $L^l$. Specifically, $L^l$ is super-resolved at $\log_{2} \alpha$ levels. 
At each level, the LR input goes through a group of spatial-angular separable convolutional (\textit{SAS-conv}) layers and a learnable transposed convolutional layers, and a residual map for the up-sampled image is finally learned using the features extracted by prior convolutions from both $L^l$ and $I^h$.

Different from single images, views in an LF image are not independent but have an implicit relation, i.e., LF structure. Specifically, under the Lambertian assumption and in the absence of occlusions, such a relation can be expressed as 
\begin{equation}
        L^l(x,y,s,t) =
        L^l(x+d \Delta s, y + d \Delta t, s + \Delta s, t+ \Delta t ), \label{eq:lfstructure}
\end{equation}
where $d$ is the disparity of the pixel located at $(x,y,s,t)$. Such a relation yields the complementary information between views (i.e., details absent at a certain view may be present in another one), which may be beneficial to the reconstruction quality. Therefore, to take advantage of this valuable information in $L^l$, we use the computationally efficient \textit{SAS-conv} \cite{lfssr2018separable} at each level to achieve the sequential feature extraction, in which 2-D convolutional layers are performed alternatively on the spatial and angular domain. The features extracted by \textit{SAS-conv} at each level will be fed into two paths: one path will be combined with the feature extracted from $I^h$ for residual prediction at the current level; the other path will be fed as input for feature extraction at the next level.
In addition, the high-capacity of a CNN model has the potential ability to handle the non-Lambertian surfaces/reflectance always occurring in practice, under which the relation in Eq. \ref{eq:lfstructure} no longer holds.

\textbf{The Down-sampling Branch.} As mentioned before, $I^h$ embeds rich information of the scene, i.e., high-frequency information and delicate textures.  This branch aims to borrow this property to enhance the learning ability of the up-sampling branch. In this branch, we progressively extract features from $I^h$ along an orientation opposite to the up-sampling branch. Specifically, at the $k$-th level ($k=0, \cdots, log_2 \alpha-1$), $I^h$ is first down-sampled to the size of $2^{-k}\alpha W \times 2^{-k}\alpha H$, and then feature maps with the same size are extracted by sequential convolutional layers. The feature maps extracted from $I^h$ at the $k$-th level are concatenated  to those from $L^l$ at the ($\log_2 \alpha-k$) level along the feature channel. The combined feature maps are further fed into the residual learning component of the up-sampling branch to reconstruct the intermediate HR LF image. 

The SR-Net is trained by minimizing the absolute error between the output denoted as $\tilde{L}_s^h$ (i.e., an intermediate HR LF images) and the ground-truth HR LF images:
\begin{equation}
    \begin{aligned}
        \ell_s^p = \| L^h - \tilde{L}_s^h  \|_1.
    \end{aligned}
\end{equation}

\textbf{Remark.} This module relies on the powerful modelling capacity of the deep CNN to super-resolve $L^l$ for an intermediate HR LF image. By progressively combining features extracted from $L^l$ and $I^h$ for the learning of HR residuals, it is expected that the SR-Net can reconstruct the HR LF image as well as possible. However, its output still suffers from blurry effects, caused by convolution and $\ell_1$ loss \cite{video2015beyondMSE,sisr2016perceptual}, although $I^h$ contains the high-frequency information of the scene. See the analysis in Sec. \ref{sec:ablation} and Figure \ref{fig:ablation}. In other words, the high-frequency information embedded in $I^h$ cannot be very effectively propagated to the output of the SR-Net. To this end, we further develop the following Warp-Net.

\subsection{Warp-Net}
As illustrated in Figure \ref{fig:subnets}, there are two sub-phases involved in this module, i.e.,  disparity estimation and inverse warping. The Warp-Net first learns an HR disparity map for each view by exploring the unique LF structure of $L^l$, which is further used to inversely warp $I^h$, leading to another intermediate HR LF image as well as its attention map.

\textbf{Disparity Estimation.} 
In this phase, we estimate the 4-D disparity map from horizontal and vertical stacks via a fully convolutional network.
Specifically, we first re-organize views of $L^l$ to construct $N$ horizontal and $M$ vertical stacks. The $i$-th (resp. $j$-th) horizontal (resp. vertical) stack contains the $i$-th row ($j$-th column) of the view array ($1\leq i\leq N$, $1\leq j\leq M$). A series of convolutional layers are applied on each stack, and those stacks corresponding to the same orientation share the weights of the convolutional kernels. The output of the final layer has the size of $W\times H\times M$ and consists of the feature maps for all views in each stack. By simply collecting $N$ horizontal feature maps, we obtain the 4-D feature map denoted as $D_h$, which only considers the horizontal structure in $L^l$. Likewise, the vertical feature map $D_v$ is computed from the vertical stacks. Then the final 4-D LR disparity map $D^l$ is generated by weighted averaging the horizontal and vertical feature maps, i.e.,
\begin{equation}
    \begin{aligned}
        D^l =  w_1  D_h + w_2 D_v,
    \end{aligned}
\end{equation}
where the weights $w_1$ and $w_2$ are adaptively learned through a convolutional layer with a kernel of size $1\times 1$.  Afterwards, $D^l$ is up-sampled with learnable transposed convolutional layers, producing the HR 4-D disparity map denoted as $D^h$.

\textbf{Inverse Warping.} Based on $D^h$, another intermediate HR LF image can be synthesized by inversely warping $I^h$ to each viewpoint. To make this module to be end-to-end trainable, we employ the differentiable bicubic interpolation \cite{STN2015} to realize the process of inverse warping.

To train the Warp-Net, we minimize the absolution error between the synthesized HR LF image $\tilde{L}_w^h$ and its ground-truth, i.e.,
\begin{equation}
    \begin{aligned}
        \ell^p_w = \| L^h - \tilde{L}_w^h  \| _1.
    \end{aligned}
\end{equation}

\textbf{Remark}. By reusing pixels from $I^h$, we expect the high-frequency details of the scene that are challenging to predict can be directly transferred from $I^h$ to each view of $\tilde{L}_w^h$. For example, for regions with continuous depths and complicated textures, Warp-Net performs quite well.  See the visual results in Figure \ref{fig:ablation}. However, $\tilde{L}_w^h$ inevitably has distortion caused by inaccurate disparity estimations or occlusions. Specifically, it is difficult to obtain accurate disparities without the ground-truth disparities for supervision, especially in challenging regions, such as textureless regions. Such inaccurate disparities will warp pixels of $I^h$ to wrong positions, resulting in distortion. Second, pixels observed in views of $L^l$ but occluded in $I^h$ will be occupied by the occluder after warping, causing error. Interestingly, the SR-Net suffers less from the distortion induced by these two factors. For example, the textureless regions, where the disparities cannot be accurately estimated, correspond to low-frequency contents, which can be relatively easily predicted by the SR-Net. Besides, the powerful regression ability of the SR-Net can predict the occluded pixels to some extent \cite{lfasr2016siggraph}.

    \begin{table*}[t]
    \centering
    \caption{Quantitative comparisons of the proposed approach with the state-of-the-art ones over the \textit{HCI} dataset. The best results are bolded, and the second best ones are highlighted with underlines.
    }
    \label{tab:quanHCI}
    \begin{tabular}{c|c|c c c c c c|c}
    \toprule
    LF image & Scale & Bicubic & PaSR \cite{lfhybrid2014iccp} & CrossNet \cite{lfhybrid2018crossnet} & SAS-conv \cite{lfssr2018separable} & M-VDSR\cite{lfssr2018separable} & M-VDSR-H & \textbf{Ours}\\
    \midrule
    Bedroom & $4\times$ & 30.86/0.897 & 34.67/0.951 & \underline{38.12/0.978} &  33.98/0.948 & 32.80/0.935 & 37.19/0.974 & \textbf{39.40/0.983}\\
    Boardgames & $4\times$ & 27.66/0.877 & 34.82/0.978 & \underline{39.88/0.992} & 33.51/0.962 & 30.45/0.929 & 36.64/0.980 & \textbf{41.65/0.993}\\
    Sideboard & $4\times$ & 23.93/0.739 & 27.05/0.874 & 30.15/\underline{0.940} & 27.98/0.898 & 26.19/0.847 & \underline{30.31}/0.936 & \textbf{33.56/0.970}\\
    Town  & $4\times$ & 28.49/0.869 & 31.64/0.930 & \underline{37.50/0.982} & 32.14/0.933 & 30.99/0.919 & 36.24/0.974 & \textbf{40.15/0.988}\\
    \midrule 
    Avg. & $4\times$ & 27.74/0.846 & 32.05/0.933 & \underline{36.41/0.973} & 31.90/0.935 & 30.11/0.908 & 35.10/0.966 & \textbf{38.69/0.983}\\
    \midrule
    Bedroom & $8\times$ & 28.29/0.844 & 33.39/0.937 & \underline{36.37/0.969} & 30.55/0.906 & 29.19/0.889 & 34.06/0.954 & \textbf{37.35/0.977} \\
    Boardgames & $8\times$ & 24.37/0.781 & 32.01/0.956 & \underline{35.99/0.983} & 27.83/0.883 & 25.53/0.832 & 30.41/0.942 & \textbf{37.84/0.987} \\
    Sideboard & $8\times$ & 21.11/0.587 & 24.58/0.768 & \underline{26.46/0.870} & 22.99/0.746 & 22.00/0.711 & 25.92/0.858 & \textbf{29.27/0.932} \\
    Town & $8\times$ & 25.59/0.794 & 29.73/0.904 & \underline{34.05/0.962} & 28.13/0.880 & 27.06/0.865 & 32.09/0.945 & \textbf{36.09/0.974} \\
    \midrule
    Avg. & $8\times$ & 24.84/0.752 & 29.93/0.891 & \underline{33.22/0.946} & 27.37/0.854 & 25.94/0.824 & 30.63/0.925 & \textbf{35.14/0.968} \\
    \bottomrule
    \end{tabular}
    \end{table*}

    \begin{table*}[t]
    \centering
    \caption{Quantitative comparisons of the proposed approach with the state-of-the-art ones over the \textit{Lytro} dataset. Here, the average values of the PSNR/SSIM over 20 LF images are reported. 
    }
    \label{tab:quanLytro}
    \begin{tabular}{c|c c c c c c|c}
    \toprule
    Scale & Bicubic & PaSR \cite{lfhybrid2014iccp} & CrossNet \cite{lfhybrid2018crossnet} & SAS-conv \cite{lfssr2018separable} & M-VDSR\cite{lfssr2018separable} & M-VDSR-H & \textbf{Ours} \\
    \midrule
    $4\times$ & 29.08/0.888 & 34.01/0.963 & 38.47/\underline{0.986}  & 33.02/0.948  & 32.41/0.943 & \underline{39.10}/0.985  & \textbf{40.22/0.988}\\
    \hline 
    $8\times$ & 26.18/0.812 & 32.39/0.951 & \underline{37.40/0.983} & 28.22/0.881 & 27.70/0.871 & 35.91/0.974  & \textbf{38.26/0.983}\\
    \bottomrule
    \end{tabular}
    \end{table*}


\subsection{Attention-Guided Fusion}
As mentioned before, the SR-Net is capable of predicting the overall content of an HR LF image but fails to recover its delicate textures and sharp edges. Warp-Net is able to propagate the high-frequency information to all views but suffers from the distortion caused by occlusions and inaccurate disparity estimation. Fortunately, their advantages are complementary to each other. Therefore, a  HR LF image can be finally reconstructed by adaptively fusing $\tilde{L}_s^h$ and $\tilde{L}_w^h$, in which their advantages are leveraged. And such an adaptive fusion process is achieved under the guidance of their own pixel-wise attention maps. 

Both attention maps are learned from the features extracted by SR-Net and Warp-Net. In SR-Net, an additional layer parallel to the output layer at the last level is  used to generate the  4-D attention map denoted as $C_s$. The loss function for training this layer is defined as:
\begin{equation}
\setlength{\abovedisplayskip}{3pt}
\setlength{\belowdisplayskip}{3pt}
    \begin{aligned}
        \ell^c_s = \frac{1}{P}\sum_{x,y,s,t} (l^h -\tilde{l}_s^h )^2  \cdot \frac{c_s}{\| C_s \|_2},
    \end{aligned}
\end{equation}
where $P$ is the total number of pixels. $l^h$, $\tilde{l}_s^h$ and $c_s$ are the pixels at $(x,y,s,t)$ in $L^h$, $\tilde{l}_s^h$ and $C_s$, respectively. In Warp-Net, similar to the disparity estimation, the network first produces a horizontal and a vertical attention  maps, which are then merged to predict the one corresponding to $\tilde{L}_w^h$, denoted as $C_w$. The up-sampling operation is also performed by transposed convolutional layers. The objective $\ell^c_w$ is defined the same way as $\ell^c_s$.

To produce the final reconstruction $\tilde{L}^h$, we first apply a Softmax normalization across $C_s$ and $C_w$, generating $\widetilde{C}_s$ and  $\widetilde{C}_w$, then weighted sum $\tilde{L}_s^h$ and $\tilde{L}_w^h$:
\begin{equation}
\setlength{\abovedisplayskip}{3pt}
\setlength{\belowdisplayskip}{3pt}
    \begin{aligned}
     \tilde{L}^h = \tilde{L}_s^h \odot  \widetilde{C}_s + \tilde{L}_w^h \odot \widetilde{C}_w,
    \end{aligned}
\end{equation}
where $\odot$ is the element-wise multiplication operator. Such an adaptive fusion process is trained under the supervision of minimizing the $\ell_1$ distance between the final reconstructed HR LF image and  the ground truth one:
\begin{equation}
\setlength{\abovedisplayskip}{3pt}
\setlength{\belowdisplayskip}{3pt}
    \begin{aligned}
        \ell^p = \| L^h - \tilde{L}^h\|_1.
    \end{aligned}
\end{equation}
Combining all modules, the whole network is trained end-to-end with the following loss function:
\begin{equation}
\setlength{\abovedisplayskip}{3pt}
\setlength{\belowdisplayskip}{3pt}
    \begin{aligned}
        \ell =  \lambda_1 \ell^p +\lambda_2 \ell^p_s +\lambda_3 \ell^c_s + \lambda_4  \ell^p_w + \lambda_5 \ell^c_w,
    \end{aligned}
\end{equation}
where all values of the weight factors $\lambda_1$, $\cdots$, $\lambda_5$ are empirically set to 1. 

    \begin{figure*}[t]
    \begin{center}
    \includegraphics[width=\linewidth]{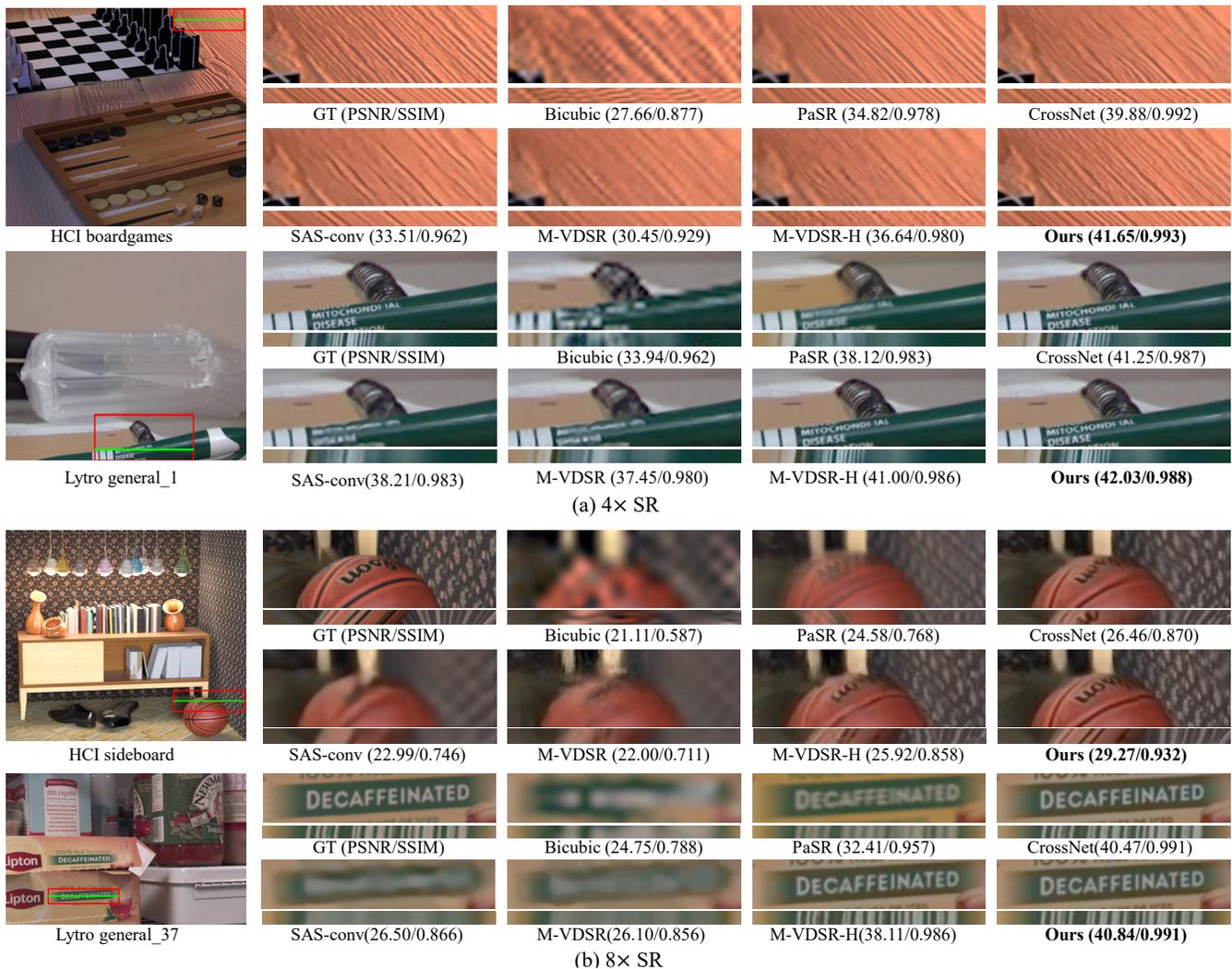}
    \end{center}
      \caption{Visual comparisons of different methods on \textit{HCI}  and \textit{Lytro}  datasets. For each algorithm, we provide the zoom-in images of the red block and EPIs constructed at the green line. The upper and bottom parts correspond to the 4$\times$ and 8$\times$ reconstruction tasks, respectively.  
    }
    \label{fig:visual}
    \end{figure*}

\section{Experiments}
\subsection{LF Data Augmentation}
\par 
Commonly used methods for data augmentation, including image rotation and flip, are not suitable for LF data. If we apply these transformations on each SAI separately, the LF structure in Eq. \ref{eq:lfstructure} would be destroyed. For example, 
applying the flip operation along the $y$ dimension, we have 
\begin{equation}
    \begin{aligned}
        &L(x,H-y,s,t) \\ &=  L(x+d\Delta s,H-( y + d\Delta t),  s + \Delta s,  t + \Delta t )\\
        &= L(x+d \Delta s,(H- y) - d\Delta t, s + \Delta s, t+ \Delta t ).
    \end{aligned}
\end{equation}
Then the corresponding pixel of the point $L(x,H-y,s,t)$ in the $(s+\Delta s, t + \Delta t)$-th SAI cannot be found,  according to the original relationship in Eq. \ref{eq:lfstructure}. 
Therefore, we propose to apply these image geometric augmentation methods on both angular and spatial dimensions. With our new strategy, taking the flip augmentation along the $y$ dimension as an example again, we have
\begin{equation}
    \begin{aligned}
        &L(x,H-y,s,N-t) \\ 
        &=L(x+d\Delta s,H-( y + d\Delta t),  s + \Delta s,  N- (t+ \Delta t))  \\
        &= L(x+d \Delta s,(H- y) - d\Delta t, s + \Delta s, (N-t)- \Delta t ), 
    \end{aligned}
\end{equation}
where the LF structure described in Eq. \ref{eq:lfstructure} is still satisfied.

\subsection{Datasets and Training Details}
   Following the existing methods \cite{lfhybrid2017ring,lfhybrid2016splitter,lfhybrid2018tci}, we simulated the data captured by the hybrid camera system via down-sampling off-center views of an HR LF image, only keeping the resolution of the central view.

    Two LF benchmark datasets were used to construct the training and test datasets,
     i.e. \textit{HCI} and \textit{Lytro}.
    \textit{HCI} is the 4-D LF benchmark \cite{lfdataset2016hci},
    and \textit{Lytro} consists of LF images captured with a Lytro ILLum camera from Stanford Lytro LF Archive \cite{lfdataset2016stanford} and the dataset provided by \cite{lfasr2016siggraph}.
    4 LF images from  \textit{HCI} and 20 LF images from \textit{Lytro} were used for testing, and the rest were used for training.
 
    The LF images were converted to YUV color space, and only the Y components were used for training and quantitative evaluation. When generating visual results, the U and V components were up-sampled using bicubic interpolation. We used Adam as the optimizer. The learning rate was initialized as $1e-4$ and $1e-5$ for $4\times$ and $8\times$ reconstruction, respectively, and reduced by a half when the loss stops decreasing. 

  \begin{table*}[t]
    \centering
    \caption{Quantitative comparisons of the proposed approach with state-of-the-art methods for LF spatial SR from hybrid inputs.
    }
    \label{tab:tci}
    \resizebox{\textwidth}{!}{
    \begin{tabular}{c c|c c c c c c c c c c }
    \toprule
    Scale & Method & Buddha & MonasRoom & StillLife & TarotCards & LegoKnights & Flowers & Amethyst & Plants2 & Leaves & Reflective29  \\
    \midrule
    \multirow{3}{*}{$4\times$} & PaSR \cite{lfhybrid2014iccp} & 32.03 & 38.52 & 25.47 & 27.47 & 31.50 & 33.04 & 34.83 & 31.79 & 31.17 & 29.01 \\
    ~ & HCSR \cite{lfhybrid2018tci} & 35.74 & 40.03 & 31.02 & \textbf{33.36} & \textbf{36.06} & 35.07 & 36.00 & 33.77 & 34.02 & 36.56 \\
    ~ & \textbf{Ours} & \textbf{43.43} & \textbf{44.34} & \textbf{31.15} & 32.16 & 32.39 & \textbf{36.92} & \textbf{40.68} & \textbf{42.24} & \textbf{37.49} & \textbf{45.08}\\
    \midrule
    \multirow{3}{*}{$8\times$} & PaSR \cite{lfhybrid2014iccp} & 27.79 & 34.21 & 23.54 & 27.73 & 29.06 & 30.49 & 31.62 & 27.38 & 23.99 & 27.20 \\
    ~ & HCSR \cite{lfhybrid2018tci} & 32.66 & 36.28 & \textbf{29.42} & \textbf{30.30} & \textbf{31.55} & 33.16  & 33.31 & 31.93 & 27.08 & 34.23 \\
    ~ & \textbf{Ours} & \textbf{40.91} & \textbf{39.39} & 28.42 & 29.22 & 29.73 & \textbf{35.10} & \textbf{37.48} & \textbf{37.82} & \textbf{31.31} & \textbf{42.23}\\
    \bottomrule
    \end{tabular}
    }
    \end{table*}

\subsection{Comparison with State-of-the-Art Methods}
We compared the proposed approach with state-of-the-art methods, including \textit{PaSR} \cite{lfhybrid2014iccp}, \textit{CrossNet} \cite{lfhybrid2018crossnet}, \textit{SAS-conv}  \cite{lfssr2018separable} and \textit{M-VDSR} \cite{lfssr2018separable}.
\textit{M-VDSR} was developed by modifying \textit{VDSR} \cite{sisr2016vdsr} to adapt to LF data, in which all SAIs of an LF image are stacked along the feature channel and then fed into the network at the same time.
Additionally, based on \textit{M-VDSR}, we developed another baseline to handle a hybrid input, namely \textit{M-VDSR-H}, in which the features from the LR LF image and HR central view are combined and fed into the \textit{VDSR} network.
Note that all the learning-based methods were re-trained over our training datasets for fair comparisons.

\textbf{Comparison of quantitative results.}
We used PSNR and SSIM to quantitatively measure the quality of the reconstructed HR LF images by different methods, and the corresponding results are listed in Tables \ref{tab:quanHCI} and \ref{tab:quanLytro}.  
 We can observe that:
\begin{itemize}  
\item methods with a hybrid input, i.e., \textit{PaSR}, \textit{CrossNet}, \textit{M-VDSR-H} and \textit{Ours}, significantly outperform those with only an LR LF input, i.e., \textit{SAS-conv} and \textit{M-VDSR}, which indicates that the extra HR view indeed makes contribution by providing more high-frequency information about the scene, and the four algorithms have the ability to take advantage of such valuable information to some extent. Also, this observation validates the potential of the hybrid LF imaging; 
\item among methods with a hybrid input, the non-learning based method \textit{PaSR} is inferior to others, indicating that an simple model with a small capacity is not enough to model the intricate relations contained in the hybrid input,  while learning-based methods, including \textit{CrossNet}, \textit{M-VDSR-H} and \textit{Ours}, have much larger capacities; and
\item our approach achieves the highest PSNR/SSIM at both datasets and scales, which can exceed the second best methods (i.e., \textit{CrossNet} or \textit{M-VDSR-H}) by more than 2 dB, demonstrating the great advantage of our method. Specifically, \textit{M-VDSR-H} simply concatenates the HR view to the LR LF image in feature space, making it difficult to model the geometric relationship between them. \textit{CrossNet} handles SAIs in an LF image independently, so it cannot make use of the valuable complementary information among SAIs and fails to preserve the LF structure. In contrast, our method is able to explicitly characterize and explore the complicated, multi-dimensional, and cross-domain relations of the hybrid input, leading to superior performance. 
\end{itemize}

\textbf{Comparison of visual results.}
We visually compared different methods for $4\times$ and $8\times$ reconstruction in Fig. \ref{fig:visual}. These results further demonstrate the significant advantages of the proposed approaches over the state-of-the-art ones, i.e., our approach can reconstruct sharper edges and clearer scenes, which are closer to the ground-truth ones. Particularly, for $8\times$ reconstruction, it is very difficult to recover the details without the guidance of an HR view. From the bottom part of Fig. \ref{fig:visual}, it can be seen that the patterns in the results of \textit{SAS-conv} and \textit{M-VDSR}  are seriously distorted. In contrast, \textit{CrossNet}, \textit{M-VDSR-H} and \textit{Ours} accept less influence of the scale increasing and can still produce acceptable results. Moreover,  our algorithm successfully preserves the high-frequency details and reconstructs sharper images, which is obvious in \textit{HCI sideboard}.

    \begin{table}[t]
    \centering
    \caption{Effectiveness verification of the fusion component in our approach. We compare the reconstruction quality of the final output with intermediate predictions under the 8$\times$ reconstruction task. The top two rows show the average PSNR/SSIM over two datasets, and the rest rows show the comparisons on several LF images.}
    \label{tab:ablation}
    \begin{tabular}{c|c c c}
    \toprule
    LF image & SR-Net & WarpNet  & Final\\
    \midrule
    HCI & 34.80/0.966 & 32.90/0.955 & \textbf{35.14}/\textbf{0.968}\\
    Lytro & 38.07/0.982 & 35.46/0.974 & \textbf{38.26}/\textbf{0.983}\\
    \midrule
    Boardgames  & 36.99/0.984  & 35.09/0.980 & \textbf{37.84}/\textbf{0.987} \\
    Town   & 35.72/0.972 &  33.51/0.964 &  \textbf{36.09}/\textbf{0.974} \\
    General8 & 46.87/0.995 & 44.85/0.994 & \textbf{47.11}/\textbf{0.996}\\
    General19 & 35.53/0.977 & 34.66/0.976 & \textbf{35.82}/\textbf{0.980}\\
    General45 & 40.67/0.986 & 38.45/0.978 & \textbf{40.93}/\textbf{0.986}\\
    \bottomrule
    \end{tabular}
    \end{table}
    
    \begin{figure}[t]
    \begin{center}
    \includegraphics[width=\linewidth]{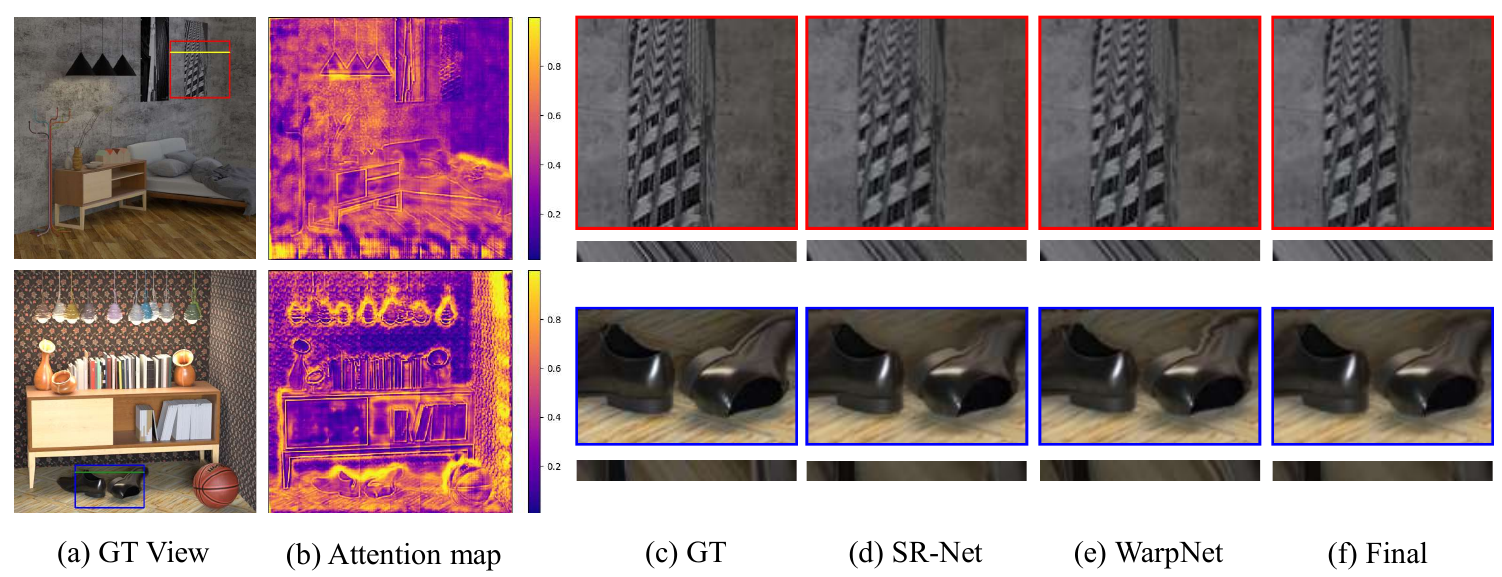}
    \end{center}
      \caption{Visual comparisons of intermediate predictions: (a) ground-truth (GT) SAIs, (b) attention maps by the SR-Net, where a larger value means higher confidence, and zoom-in blocks extracted from the (c) GT SAIs, (d) predictions by the SR-Net, (e) predictions by the WarpNet, and (f) final outputs. \textit{Red} frames in the first row highlight the advantages of WarpNet and \textit{Blue} frames in the second row highlight the advantages of SR-Net.}
    \label{fig:ablation}
    \end{figure}

\textbf{Comparison of the LF parallax structure.}
The most valuable information of LF data is the LF structure in Eq. \ref{eq:lfstructure}, which implicitly represents the geometry of the scene/object. Here, we evaluated the LF structure of the reconstructed HR LF images by different methods both qualitatively and quantitatively.  Comparing the 2-D epipolar plane image (EPI) is a straightforward way to evaluate the LF structure qualitatively. 
In the EPI of a real LF image, the projections of a single scene point observed in different SAIs construct a straight line. Therefore, we present EPIs constructed from the predictions of different algorithms for comparison in Fig. \ref{fig:visual}. It can be seen that the EPIs of our algorithm have clearer line texture and more accurate slops, which demonstrates that our network preserves the LF structure better than others. For example, in \textit{Lytro general$\_$37}, although \textit{CrossNet} and \textit{M-VDSR-H} can reconstruct the letters as clear as \textit{Ours}, they fail to reconstruct the correct LF structure, which is reflected by the non-straight EPI lines.

\begin{figure}[t]
    \centering
    \subfigure{
    \includegraphics[width=0.4\linewidth]{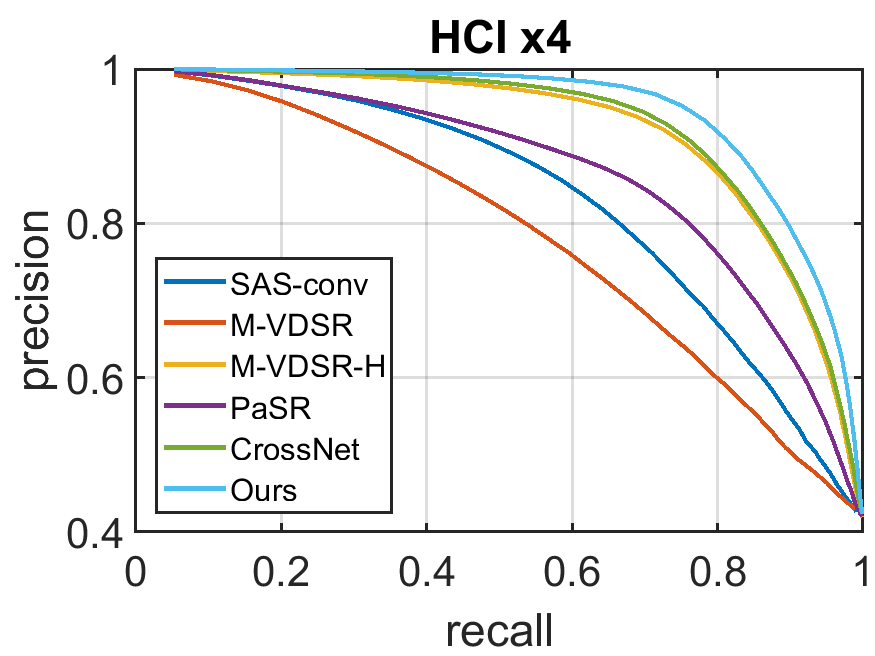}
    }
    \subfigure{
    \includegraphics[width=0.4\linewidth]{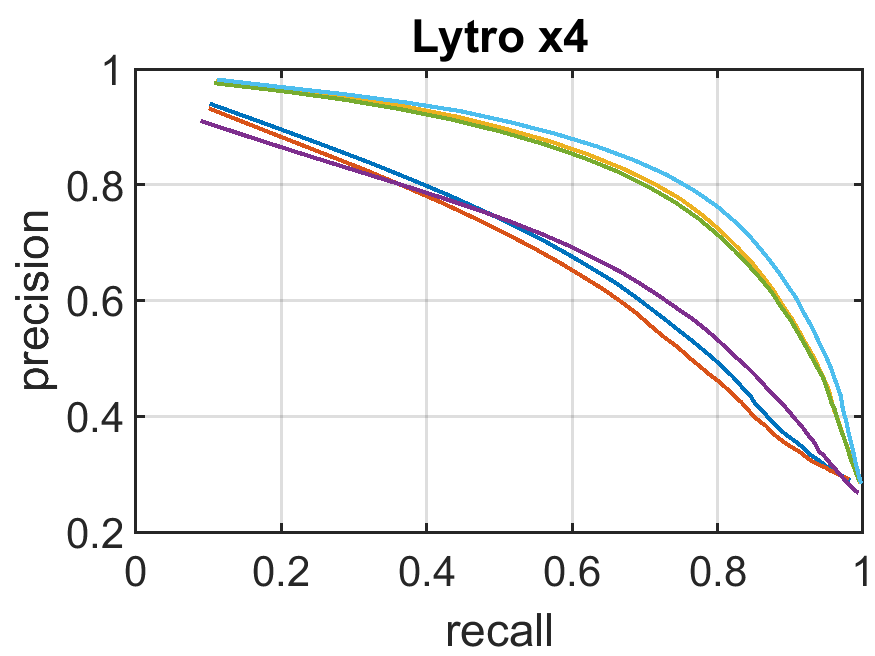}
    }\\
    \subfigure{
    \includegraphics[width=0.4\linewidth]{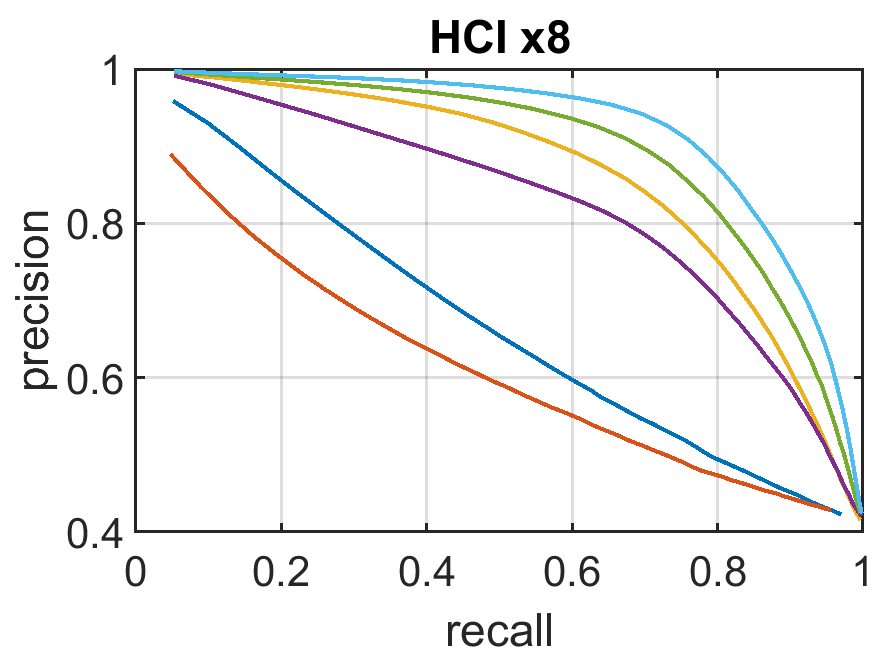}
    }
    \subfigure{
    \includegraphics[width=0.4\linewidth]{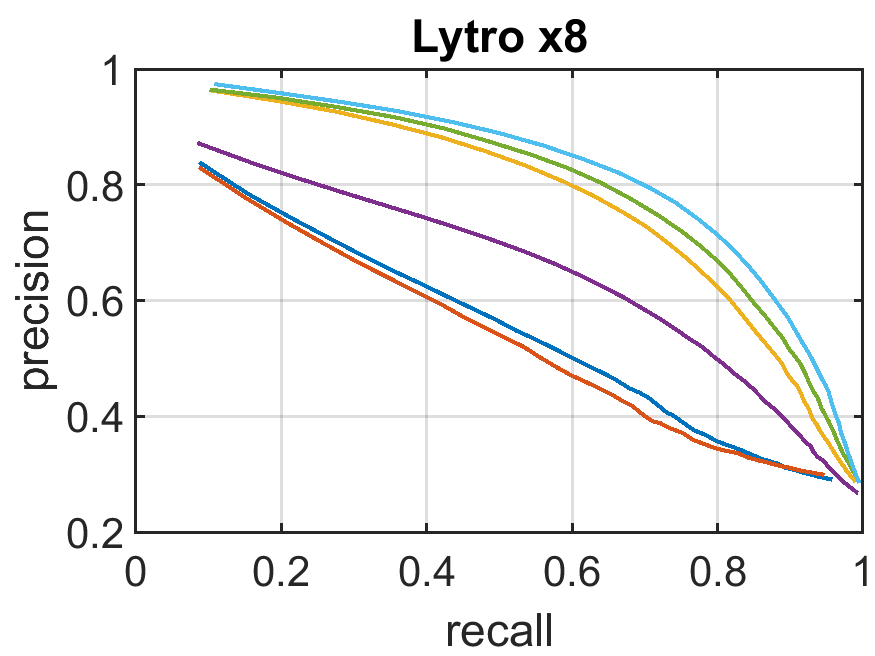}
    }
    \caption{Parallax content PR curves for different methods. The averages across each dataset are presented. All subfigures share the same legend  shown in the first one.
    }
    \label{fig:prcurve}
    \end{figure}

We also quantitatively evaluated the LF structure by using the LF parallax edge precision-recall (PR) curves \cite{lfapp2018denoising}, and Fig. \ref{fig:prcurve} shows the corresponding results, where it can be seen that the PR curves by our methods are closer to the top-right corner than the others, demonstrating that our method preserves the LF parallax best.

\textbf{Comparison with \textit{HCSR}}.
\textit{HCSR} \cite{lfhybrid2018tci} is a state-of-the-art method aiming at super-resolving LR LF images that are arranged around an HR reference central view, which is exactly the same task as ours. 
It is a traditional step-wise method which involves two modules, i.e., single view super-resolution and warping-based difference compensation.
As the code of \textit{HCSR} is not publicly available, we conducted experiments on the same data provided in \cite{lfhybrid2018tci} and compared with the results of \textit{PaSR} and \textit{HCSR} reported in \cite{lfhybrid2018tci}. As shown in Table \ref{tab:tci}, our method achieves significant improvement on the PSNR values of most scenes. \textit{HCSR} outperforms our method on 2 images with baselines much larger than those of our training datasets. 


\textbf{Efficiency.} 
We also compared the computational complexities of different methods by measuring the running time (in second) of the testing phase, and Table \ref{tab:time} lists the results. All methods were tested on a desktop with Intel CPU i7-7700@3.60GHz, 64 GB RAM and NVIDIA GeForce GTX 1080 Ti. From Table \ref{tab:time}, it can be observed that our approach is much faster than \textit{CrossNet} and \textit{PaSR}, and slightly slower than the others. But taking the trade-off between the computational complexity and reconstruction quality, we believe our method is the best one.

\subsection{Ablation Study}
\label{sec:ablation}
To validate the effectiveness of the fusion component,
we compared the reconstruction quality of the intermediate predictions by SR-Net and WarpNet and the final output under the 8$\times$ reconstruction over \textit{HCI}.
From Table \ref{tab:ablation}, it can be seen that the PSNR/SSIM values of the final output are higher than those of the intermediate predictions, and especially on the \textit{Boardgames} image, the PSNR is improved more than 0.8 dB and 2.7 dB with respect to the SR-Net and WarpNet, respectively, demonstrating the effectiveness of the fusion component. To further investigate the difference between SR-Net and WarpNet as well as their contributions to the final output, we visually compared the intermediate predictions and the corresponding attention maps in Fig. \ref{fig:ablation}. For plain areas (\textit{red} frames in the first row), the prediction of SR-Net is blurred while that of WarpNet has sharper edges.
The attention map also shows that WarpNet has higher confidence in these areas.
For areas with discontiguous depth (\textit{blue} frames in the second row), the prediction of WarpNet has distortion while that of SR-Net maintains the content and has higher confidence. Therefore, we can conclude that the fusion is indeed able to leverage the advantages of these two modules. In addition, EPIs in Fig. \ref{fig:ablation} demonstrate the improvement on the LF structure preservation after the fusion. 

    \begin{table}[t]
    \centering
    \caption{Comparisons of the average running time (in second) of different algorithms for reconstructing an HR LF image.}
    \label{tab:time}
    \resizebox{ \linewidth}{!}{
    \begin{tabular}{c|c c c c c c}
    \toprule
      & PaSR & CrossNet & SAS-conv  & M-VDSR & M-VDSR-H & Ours\\
    \midrule
    HCI $4\times$ & 4556.61 &  15.30 & 3.73 & 0.29 & 2.51 &  6.31 \\
    HCI $8\times$ & 1191.30 &  15.32 & 3.89 & 0.30 & 2.51 &  6.54 \\
    Lytro $4\times$ & 2152.14 &  7.72 & 1.37 & 0.13 & 1.23 &  2.80 \\
    Lytro $8\times$ & 567.67 &  7.73 & 1.43 & 0.13 & 1.23 &  2.88 \\
    \bottomrule
    \end{tabular}
    }
    \end{table}

\section{Discussion}

We propose this attention-guided fusion strategy to research hybrid information, which is believed to inspire other hybrid LF processing algorithms.
We believe our proposed framework is the key element for a practical hard-ware based hybrid imaging system. The respectively advantages and error-prone pixels from different lenses have been learnt by the attention module for optimized fusion. Based on this work, when implementation details such as camera calibration and handling of arbitrary off-grid camera inputs are investigated, the real potential of a hybrid system can be practically realized.
    
In view of the impressive quality achieved in the reconstruction of a large up-sampling scale, we believe our framework could potentially decrease the cost of high-quality LF data acquisition for speeding the deployment of LF technique in practice, such as immersive communication. Under this setting, our method can also save the storage cost. In addition, our framework will be potentially beneficial to the compression of LF data directly acquired by other advanced devices, which is an emerging and high desirable issue for LF based immersive communication \cite{lfapp2019houcompression}. For example, at the server side, the spatial resolution of all SAIs except a certain one involved in an LF image can be reduced and likewise the decrease of its data size.  And the data can be recovered with our framework at the clients.

\section{Conclusion}
We have presented a novel learning-based method for reconstructing an HR LF image from a hybrid input. Owing to the elegant and innovative network architecture, which is capable of comprehensively taking advantage of the underlying properties of the input from two complementary and parallel perspectives, our method not only achieved more than 2 dB improvement in terms of the reconstruction quality and preserved the LF structure much better, but also run in an end-to-end manner and at a high speed, when compared with state-of-the-art approaches. 

Based on this work, we will continue to evaluate the proposed method over real data acquired by a  typical hybrid imaging system. Specifically, implementation details such as camera alignment is going to be carefully studied. Additionally, a more flexible framework which enables inputs from arbitrary off-grid LR camera positions is to be investigated, which could eventually release the real potential of a hybrid imaging system.


\bibliographystyle{ACM-Reference-Format}
\bibliography{reference}


\end{document}